\documentclass[10pt,twocolumn,letterpaper]{article}

\usepackage{multirow}
\usepackage{cvpr}
\usepackage{times}
\usepackage{epsfig}
\usepackage{graphicx}
\usepackage{amsmath}
\usepackage{amssymb}

\usepackage[breaklinks=true,bookmarks=false]{hyperref}

\cvprfinalcopy 

\newcommand{\norm}[1]{\left\lVert#1\right\rVert}

\def\cvprPaperID{xxxx} 

\begin{document}

\title{Texture Fuzzy Segmentation using Skew Divergence Adaptive Affinity Functions}

\author{Jos\'{e} F. S. Neto\\
Federal University at Rio Grande do Norte\\
{\tt\small netolcc06@gmail.com}
\and
Waldson P. N. Leandro\\
Federal University at Rio Grande do Norte\\
{\tt\small waldsonpatricio@gmail.com}
\and
Matheus A. Gadelha\\
University of Massachusetts Amherst\\
{\tt\small mgadelha@cs.umass.edu}
\and
Tiago S. Santos\\
Brazilian Federal Highway Police Station\\
{\tt\small souza.san@gmail.com}
\and
Bruno M. Carvalho\\
Federal University at Rio Grande do Norte\\
{\tt\small bruno@dimap.ufrn.br}
\and
Edgar Gardu\~{n}o\\
Universidad Nacional Aut\'{o}noma de M\'{e}xico\\
{\tt\small edgargar@unam.mx}
}

\maketitle

\begin{abstract}
   Digital image segmentation is the process of assigning distinct labels to different objects in a digital image, and the fuzzy segmentation algorithm has been successfully used in the segmentation of images from a wide variety of sources. However, the traditional fuzzy segmentation algorithm fails to segment objects that are characterized by textures whose patterns cannot be successfully described by simple statistics computed over a very restricted area. In this paper, we propose an extension of the fuzzy segmentation algorithm that uses adaptive textural affinity functions to perform the segmentation of such objects on bidimensional images. The adaptive affinity functions compute their appropriate neighborhood size as they compute the texture descriptors surrounding the seed spels (spatial elements), according to the characteristics of the texture being processed. The algorithm then segments the image with an appropriate neighborhood for each object. We performed experiments on mosaic images that were composed using images from the Brodatz database, and compared our results with the ones produced by a recently published texture segmentation algorithm, showing the applicability of our method.
\end{abstract}

\section{Introduction}\label{sec:Intro}

Digital image segmentation is the process of assigning distinct labels to different objects in a digital image. In order to perform object identification on digital or continuous images, humans make use of several high-level information, such as shadowing, occlusion, parallax motion, relative size of objects as well as all a priori knowledge about objects. Besides being very difficult to insert this type of reasoning into a computer program, the task of segmenting out an object from a background in an image becomes very hard for a computer when, instead of intensity values, what distinguishes the object from the background is some textural property, more so when the image is affected by noise and/or inhomogeneous illumination.

Physical textures are usually described as the feel or shape of a surface or substance, while in the areas of Visual Computing, such as Image Processing and Computer Vision, they are characterized by statistical distributions of the observed intensities or repeated arrangements of a kernel image. Haralick \cite{HARA79a} classified textures as structural or non-structural, where structural textures have a repeated pattern associated with a texture periodicity \cite{JAN98a}, and non-structural or statistical textures do not possess this characteristic and are solely described by statistical distributions.

Texture analysis can involve synthesis, segmentation or classification and it is a challenging task (due to the nature of the textures as well as the effects that changes in scale, illumination or noise may have in their perception). In this article, we are concerned with the segmentation of images formed by a composition of natural non-structural textures, by using a semi-automatic, interactive region-growing segmentation method, called \emph{fuzzy segmentation}, that is extended by the usage of adaptive fuzzy affinity functions that are appropriate for characterizing and comparing textures. These adaptive fuzzy affinity functions have scales that are related to the statistical distribution of the textural patterns they are associated with.

The interactive region-growing method used to segment the images is based on the fuzzy connectedness, a concept that has been successfully used to segment noisy images, as can be seen in \cite{CARV06a}. The original fuzzy segmentation method used here, based on the work of \cite{UDUP96a}, was first proposed in \cite{HERM01a} and improved in \cite{CARV05a}. This is a semi-automatic method that belongs to the class of region-growing segmentation algorithms, and the user interaction with the method happens in the form of selected seed spels (short for spatial elements) such as pixels in the case of 2D images. However, we also propose in this article a method for automatizing the seed selection, thus transforming the method into a fully automatic texture segmentation method.


\section{Related Work}\label{sec:RelWork}
In the literature of classification, segmentation and analysis of textures, the textures are referred as being structural (or periodic) and non-structural (or statistical) textures. There are several techniques that can be used to extract texture periodicities and determine the minimal kernel that fully characterizes a structural texture. However, in this paper, we are concerned with the much harder problem of segmenting images that contain objects which are characterized by statistical textures.

Several different techniques have been employed to perform texture segmentation. One example of these techniques is the one proposed by \cite{UNSE89a}, that extracts local texture properties using local linear transforms that were optimized for maximal texture discrimination. Then, they estimate local statistics at the output of a filter bank and generate a multi-resolution sequence using an iterative Gaussian smoothing algorithm. Scatter matrices are then evaluated to reduce the texture features to a single component that is then thresholded to produce the segmentation. A very different approach is proposed by \cite{HOFM98a} that also performs unsupervised texture segmentation, but does it by relying on statistical tests as a measure of homogeneity. The texture segmentation task is then formulated as a clustering problem and solved by computing dissimilarities using multi-scale Gabor filters.

There are several model-based approaches \cite{CHEL85a,MANJ91a,ROBE95a} that approach the texture segmentation problem by modeling the intensity field of textures as a Gauss-Markov or as a Markov random field to represent the local spatial dependencies between pixel intensities. However, these techniques are usually very computationally intensive because they require a large number of iterations to converge. Lehmann \cite{LEHM11a} proposed an alternative approach that models two-dimensional textured images as the concatenation of two one-dimensional hidden Markov auto-regressive (HMM-AR) models, one for the lines and one for the columns. Then, the segmentation is performed by an algorithm that is similar to turbo decoding in the context of error correcting codes, where the unknown parameters of the HMM-AR are estimated using the Expectation-Maximization algorithm.

\section{Fuzzy Segmentation}\label{sec:FuzzySeg}
The method of Multi-Object Fuzzy Segmentation (MOFS) proposed in \cite{HERM01b} is based on the concept of fuzzy connectedness and it is applicable to arbitrary digital spaces \cite{HERM98c} as well as to the simultaneous multiple object segmentation. Because of the general nature of the approach used, we refer to elements of the set $V$ to be segmented as \emph{spels}, which is short for \emph{spatial elements} \cite{HERM98c}. The spels can be pixels of an image  \cite{CARV05a}) but they can also be dots in the plane \cite{AHUJ82a}, or any variety of other things, such as feature vectors \cite{DUDA00a}. Thus, the theory and algorithm mentioned here can also be applied to data clustering in general \cite{JAIN99a}.

The goal is to partition $V$ into a specified number of objects, but in a fuzzy way, by assigning a spel to a particular object with a grade of membership (a number between 0 and 1) to that object. In order to do that, we assign, to every ordered pair $(c,d)$ of spels, a real number not less than 0 and not greater than 1, which is referred to as the \emph{fuzzy connectedness} of $c$ to $d$. 

In the original approach of Carvalho et al. \cite{CARV05a}, fuzzy connectedness is defined in the following general manner. We call a sequence of spels a \emph{chain}, and its \emph{links} are the ordered pairs of consecutive spels in the sequence. The strength of a link is also a fuzzy concept. We say that the $\psi$-\emph{strength} of a link is the value of a \emph{fuzzy spel affinity} function $\psi:V^{2}\rightarrow[0,1]$. A chain is formed by one or more links and the $\psi$\emph{-strength of a chain} is the $\psi$-strength of its weakest link; the $\psi$-strength of a chain with only one spel in it is $1$ by definition. 

The fuzzy connectedness from $c$ to $d$ is computed using a \emph{fuzzy connectedness} function $\mu_{\psi}:V^{2}\rightarrow[0,1]$ defined by  the $\psi$-strength of the strongest chain from $c$ to $d$. We can then define the $\psi$-\emph{connectedness map} $f$ of a set $V$ for a seed spel $o$ as the picture formed by the fuzzy connectedness values of $o$ to $c$ ($f(c)=\mu_{\psi}(o,c)$), for all $c\in V$. 

An $M$-\emph{semisegmentation} of $V$ is a function $\sigma$ that maps each $c\in V$ into an $(M+1)$-dimensional vector $\sigma ^{c}=(\sigma _{0}^{c},\sigma _{1}^{c},$ $\cdots ,$ $\sigma _{M}^{c})$, such that $\sigma _{0}$ is non-negative but not greater than 1, for each $m$ ($1\leq m\leq M$), the value of $\sigma _{m}^{c}$ is either 0 or $\sigma _{0}^{c}$, and, for at least one $m$ ($1\leq m\leq M$), $\sigma _{m}^{c}=\sigma _{0}^{c}$. Finally, an $M$-\emph{segmentation} of $V$ is an $M$-semisegmentation such that $\sigma_0^c > 0$ for every spel $c \in V$. There is an unique $M$-\emph{segmentation} asociated with a set $V$ and sets of seed spels $V_m$, for $1 \leq m \leq M$, that is n fact the segmentation of that set.

For more information related to the fuzzy segmentation algorithm, the reader should refer to \cite{CARV05a}, where the FAST-MOFS algorithm is introduced. The FAST-MOFS is a greedy algorithm that computes the grades of membership of all spels to the objects rounding them to three decimal places. This quantization of the fuzzy affinity functions allows us to use an array as a priority queue, instead of a heap, thus lowering the computational complexity of the algorithm.

\subsection{Defining Fuzzy Affinity Functions}
In the semi-automatic setting of our method, we originally specified $\psi _{m}$ and $V_{m}$ $(1\leq m\leq M)$ for such an image by clicking on some spels in the image to identify them as belonging to the $m$th object, and $V_{m}$ is formed by these points and their eight neighbors. We then defined $g_{m}$ to be the mean and $h_{m}$ to be the standard deviation of the average brightness for all edge-adjacent pairs of spels in $V_{m}$ and $a_{m}$ to be the mean and $b_{m}$ to be the standard deviation of the absolute differences of brightness for all edge-adjacent pairs of spels in $V_{m}$. Finally, we defined the affinity functions as:
\begin{equation}
\psi(c,d)= \left\{ \begin{array}{lr}  
\frac{\rho_{g_m , h_m}(g) + \rho_{a_m , b_m}(a)}{2}, & \text{if $c$ and $d$ are}\\
& \text{edge-adjacent}, \\ 
0, & \text{otherwise},
\end{array}  
\right.
\label{eq:fuzzy-affinity-function}
\end{equation}
where $g$ is the mean and $a$ is the absolute difference of the brightnesses of $c$ and $d$ and the function $\rho_{r,s}(x)$ is the probability density function of the Gaussian distribution with mean $r$ and standard deviation $s$ multiplied by a constant so that the peak value becomes $1$.

It is clear from Equation (\ref{eq:fuzzy-affinity-function}) that the first term measures locally the degree of homogeneity by computing the average of the two intensities and passing it to the fuzzy affinity function, while the second term computes the absolute difference between the two intensity values, capturing some inhomogeneity information, that can be construed as a very basic textural information. 

\subsection{Textural Fuzzy Affinity Functions}

While Equation (\ref{eq:fuzzy-affinity-function}) has been successfully used in several types of textures \cite{CARV12a}, images \cite{CARV05a,CARV06a}, and even grids \cite{CARV01a}, we detected that it cannot capture enough information about more complex textures, specially the ones whose textural characteristics can only be described at larger scales. By analyzing the results, we detected the need to encode more information about the intensity distribution into the affinity functions. The idea is to compare intensity distributions, thus providing a more powerful and robust way to describe the visual characteristics of textures. 

In the fields of Statistics and Information Theory, the well-known Kullback-Leibler (KL) divergence \cite{KULL51}, also known as relative entropy, is a non-symmetric measure of the distance or difference between two probability distributions $q$ and $r$. In Information Theory, the KL divergence of $r$ from $q$, denoted by $KLD(q||r)$, is a measure of the information lost when $r$ is used to approximate $q$, and is given by Equation \ref{eq:KL}:

 \begin{equation}
 KLD(q || r) = \sum_y q(y)(\log q(y) - \log r(y)).
 \label{eq:KL}
 \end{equation}
 
The KL divergence is in fact a tool to measure how far away two distributions are. However, the KL divergence is not defined when $r(y)=0$, and that happens quite often when we are dealing with distributions that are characterized by histograms with few counts. For example, if we are dealing with monochromatic images with $256$ gray levels and neighborhoods of size $11 \times 11$, we are characterizing a distribution by using a histogram with $256$ bins and whose total count is only $121$. In order to deal with this problem, Lee \cite{Lee99} proposed an approximation to the KL divergence, that is given by Equation \ref{eq:skew}:
\begin{equation}
SD_\alpha(q, r) = D(r || \alpha q + (1 - \alpha)r),
\label{eq:skew}
\end{equation}
for $0 \leq \alpha \leq 1 $, where $\alpha$ can be seen as a degree of confidence in the distribution $q$ and $(1 - \alpha)$ as the amount of smoothing of the distribution $q$ over $r$. This allows us to work nicely with sparse data problems, such as the low count histograms of small neighborhoods.


\subsection{Adaptive Fuzzy Affinity Functions}\label{subsec:AdapFuzzyAffFunc}
There are several techniques for detecting the window size that characterizes the texture periodicity pattern \cite{JAN98a} and even orientation \cite{LIN97a} of structural textures. However, such area is not well-defined in the case of non-structural textures, which can exhibit localized distributions of intensities in smaller scales and very different distributions observed at larger scales.

In this work, in order to include more information about textural properties in the segmentation process, we use adaptive fuzzy affinity functions whose neighborhood areas are selected according to the distribution of the intensities surrounding the selected seed spels. This selection defines the scale in which the correspondent texture will be segmented. As mentioned before, the original fuzzy affinity functions used the intensity values of two spels $c$ and $d$ to compute the strength of the link connecting $c$ to $d$. Here, in order to capture some information about the intensities in the neighborhood area (scale) defined for the texture in question, we compute the average and the standard deviation of the pairs of neighbors in the defined neighborhood. These values, when computed for the seed spels, are used to define the second term of (\ref{eq:fuzzy-affinity-function}), and during the segmentation algorithm, the values computed in the neighborhoods of spels $c$ and $d$ are used as an input to this term.

\begin{figure*}
\begin{center}
\begin{tabular}{ccc}
 \includegraphics[scale=0.25]{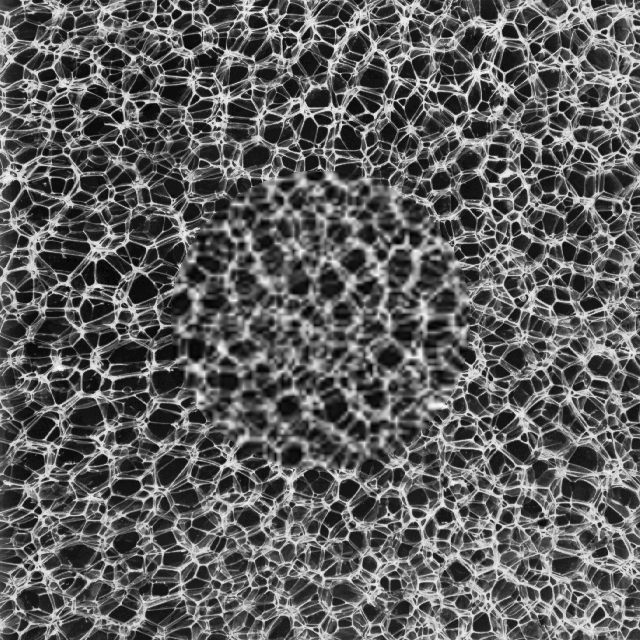}&
 \includegraphics[scale=0.25]{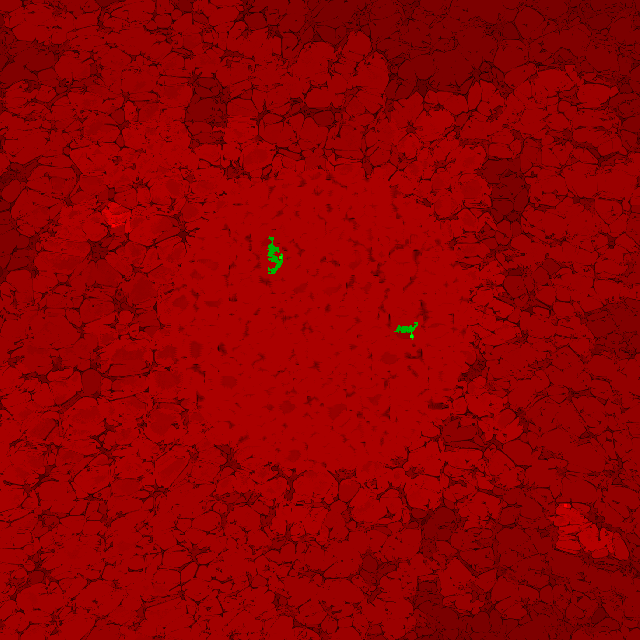}&
 \includegraphics[scale=0.25]{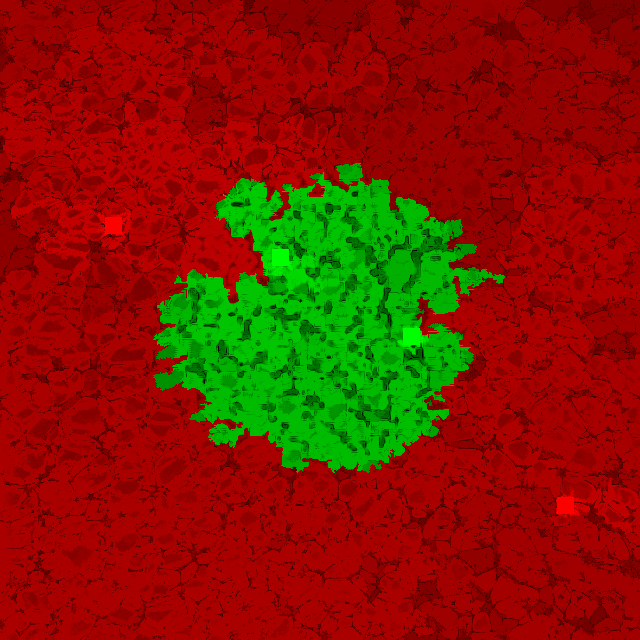}\\
(a)&(b)&(c)\\
 \includegraphics[scale=0.25]{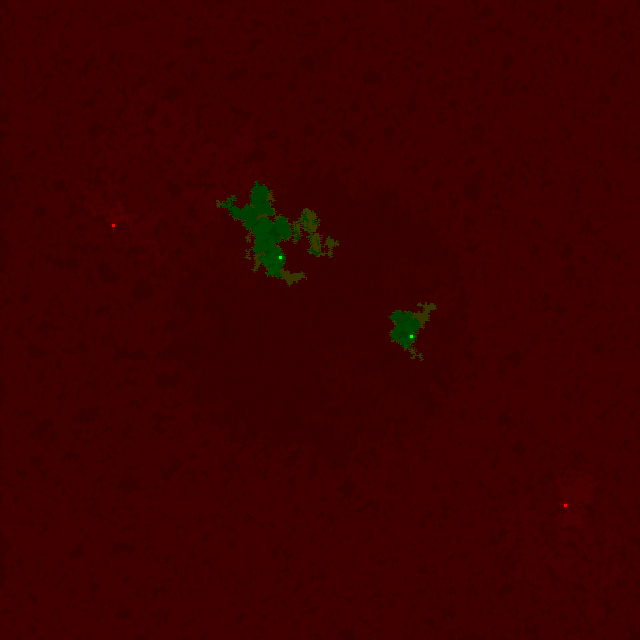}&
 \includegraphics[scale=0.25]{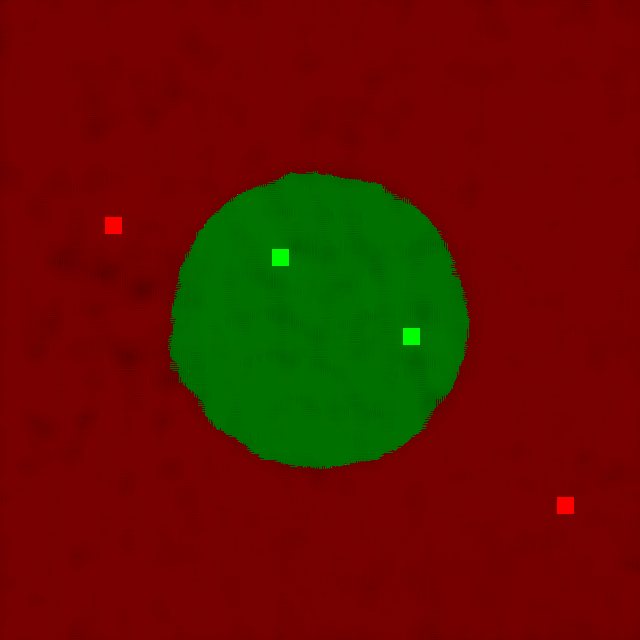}&
 \includegraphics[scale=0.25]{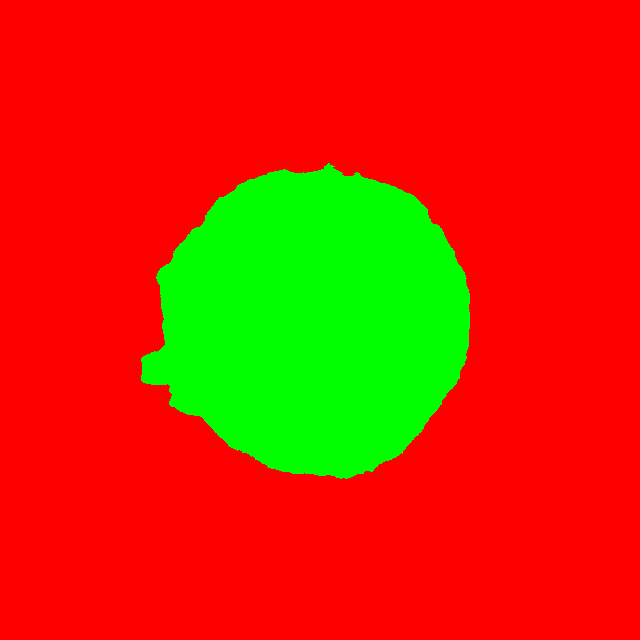}\\
(d)&(e)&(f)
\end{tabular}
\end{center}
\caption{Original Mosaic produced by combining the same texture at two different scales (a) and correspondent connectedness maps produced using the Gaussian affinity function with a $3 \times 3$ (b) and adaptive (c) neighborhoods, and the Swew Divergence affinity function with a $3 \times 3$ (d) and adaptive (e) neighborhoods, and the segmentation produced bu the method of Yuan et al. \cite{YUAN15a}.}
\label{ScalesTexturesSeg}
\end{figure*}

We find the appropriate scale for each object by computing statistics for a neighborhood of size $3 \times 3$ surrounding the seed spels, and increasing this neighborhood until the differences of the statistics collected on the neighborhoods of two consecutive sizes are smaller than predefined thresholds. We do that by computing a measure of the distance between the regions surrounding two seed spels. In the case of the Gaussian-based fuzzy affinity functions, we use a function of the differences between the averages and the standard deviations, that were empirically set to $0.06$ and $0.04$, respectively, while in the case of the Skew divergence we used as a threshold a minimum value for the divergence measure, that was empirically set to $0.8$. These thresholds were selected empirically for detecting the point when the collected statistics stabilize. We also define a maximum size for the neighborhoods. The function used to compute the statistics is generic in the sense that it can work with square, circular (digital ball), hexagonal or any other neighborhood shape, following the general definitions given in \cite{HERM98c}.





\section{Results and Discussion}\label{sec:Discussion}

In order to show the applicability of our approach, we first selected 10 textures from the Brodatz textures data set \cite{BROD66a} that exhibit features on several different scales. We selected other Brodatz textures that were not used in the segmentation experiments shown below to determine empirically values for the thresholds mentioned in Section \ref{subsec:AdapFuzzyAffFunc}, that were set to $0.06$ and $0.04$ for the mean and standard deviation, respectively.

In all the segmentation results shown here, we display the connectedness maps with color indicating the object to which the spel belongs to while the intensity is proportional to the grade of membership of that spel to the object in question, i.e., spels that appear very dark have a low grade of membership to the object to which they belong to. All experiments were run on a notebook equipped with an Intel i5 CPU running at 1.8 GHz and 6GBs of RAM.

Figure \ref{ScalesTexturesSeg} shows the segmentations of one $640 \times 640$ image that was composed by putting together two objects made out of the same texture from the Brodatz data set but on different scales in order to show the importance of the scale computation to properly characterize a texture. In this experiment, two seeds were selected for each texture, and the scale for each object was determined according to the procedure described on Section \ref{subsec:AdapFuzzyAffFunc}. We used square neighborhoods and the largest neighborhood used was $15 \times 15$. It is clear from the results that our approach had no problem in segmenting these textures when using adaptive fuzzy affinity functions. One can clearly see on Figures \ref{ScalesTexturesSeg}(b) and \ref{ScalesTexturesSeg}(d) that the neighborhood size of $3 \times 3$ was not sufficiently large to characterize the texture pattern of the objects. Not only it led to a very poor segmentation, but also to very low connectedness values (as we can see from the overall low intensity values of the map), or very low confidence on the result. It is important to point out that the colors shown on Figure \ref{ScalesTexturesSeg}(e) do not change in intensity within the same object because the method used for producing it is crisp, i.e., is not fuzzy in nature, and thus, does not compute a grade of membership for each spel.

\subsection{Automatic Segmentation}

Since our method performs semi-automatic segmentation,the quality  of the final result is strongly correlated to the initial seed positions, which are selected manually. We overcome this issue by selecting the seeds through a clustering method in a embedded space induced by the skew-divergence metric. By using this approach, the whole segmentation process can be done without relying on human selection of the seeds, just the number of classes. Our procedure is defined as follows.

We start by dividing the input image in $N \times N$ patches, where $\pi_i$ is the $i$th patch. Then, we compute the pixel intensity histogram $H_i$ for each patch. Using this histogram, we compute a pairwise distance matrix $D$ using the following equation:
\begin{equation}
D_{ij} = \frac{1}{2} \bigg( KL(H_i || H_j) + KL(H_j || H_i) \bigg) + \lambda \norm{ \mu(\pi_i) - \mu(\pi_j) }_2 ,
\end{equation}
where $\lambda$ is a user-defined weight and $\mu(\pi)$ is an operator that returns the mean coordinate of pixels in the patch $\pi$. After preliminary experiments, we decided to set $\lambda=0.5$ in all of our following experiments. The reason for adding the second term is to increase the distance between patches that are far from each other in the original image. This is particularly useful in our region-growing based method: distant pixels have a smaller probability of belonging to the same class of the original seeds. The matrix $D$ is used to create an embedded space through Multi-Dimensional Scaling (MDS) \cite{BORG05a}. 

In other words, MDS will output a set $\mathcal{P}$ in a 3-dimensional space that preserves the distances in $D$. The set $\mathcal{P}$ and the number of classes $k$ will serve as an input to the $k$-means clustering \cite{DUDA00a} procedure that will label each point $p \in \mathcal{P}$ as belonging to one class. From the MDS computation and the $k$-means, we can associate each patch $\pi$ to a class $c$. Using this information, we define $k$ disjoint sets of patches $\Pi_1, \Pi_2, ..., \Pi_k$. The centroid $\bar{\mu_c}$ of a region assigned to a class $c$ is computed as follows.

\begin{equation}
\overline{\mu_c} = \frac{1}{|\Pi_c|}\sum_{\pi \in \Pi_c} \mu(\pi) \label{eq:muc}
\end{equation}

One could simply use $\mu_c$ of Equation (\ref{eq:muc}) as a seed position, but we noticed through experimentation that sampling multiple seeds for each class is more effective than using a single seed. Thus, we select the seeds of a class $c \in [1, k]$ by sampling from a multivariate normal distribution $\mathcal{N}(\bar{\mu_c}, I)$, where $I$ is a $2\times2$ identity covariance matrix. We used 3 samples per class in all automatic MOFS experiments shown in this paper. 

\subsection{Experiments}

\setlength{\tabcolsep}{6pt}
\begin{table*}
\begin{center}
\begin{tabular}{|c|c|c|c|c|c|c|}
\hline
\multirow{3}{*}{Mosaic}& \multicolumn{2}{|c|}{MOFS Skew}& \multicolumn{4}{|c|}{\cite{YUAN15a}}  \\
\cline{4-7}
  & \multicolumn{2}{|c|}{ } & \multicolumn{2}{|c|}{$60 \times 60$} & \multicolumn{2}{|c|}{$120 \times 120$}  \\
\cline{2-7}
  & Dice & Time & Dice & Time & Dice & Time  \\
\hline
M2-1 & \textbf{0.994} & 13.197 & 0.986 & 2.832 & 0.980 & 2.965 \\
\hline
M5-1 & \textbf{0.987} & 50.53 & 0.670 & 16.22 & 0.955 & 30.45   \\
\hline
M5-2 & \textbf{0.988} & 51.97 & 0.600 & 17.89 & 0.615 & 28.39   \\
\hline
M5-3 & 0.966 & 51.36 & 0.690 & 13.61 & \textbf{0.974} & 26.29   \\
\hline
\end{tabular}
\end{center}
\caption{Accuracies of fuzzy segmentations (in percentage) produced using the Gaussian affinities and Skew affinities with a pre-defined affinity neighborhood and the automatic adaptive neighborhood.}
\label{AccuracyTable}
\end{table*}

\begin{figure}[!t] 
\centering
\includegraphics[scale=.5]{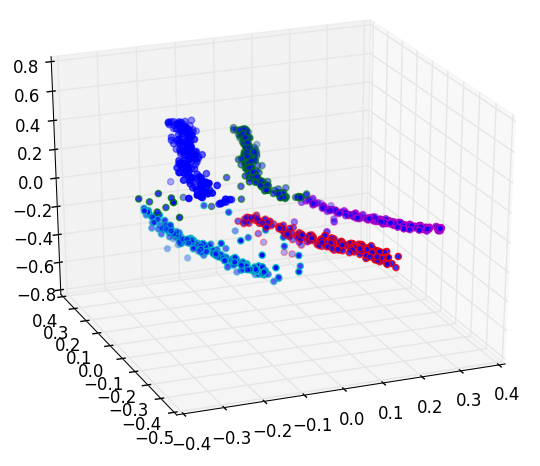}
\caption{Clustering into 5 objects of the $64 \times 64$ patch centers for Mosaic 1 (Figure \ref{MosaicFiveTextures}(a)) where can see the clear separation of the patches belonging to the different textures.}\label{fig:cluster01}
\end{figure}

\begin{figure*}
\begin{center}
\begin{tabular}{ccc}
\includegraphics[scale=0.1]{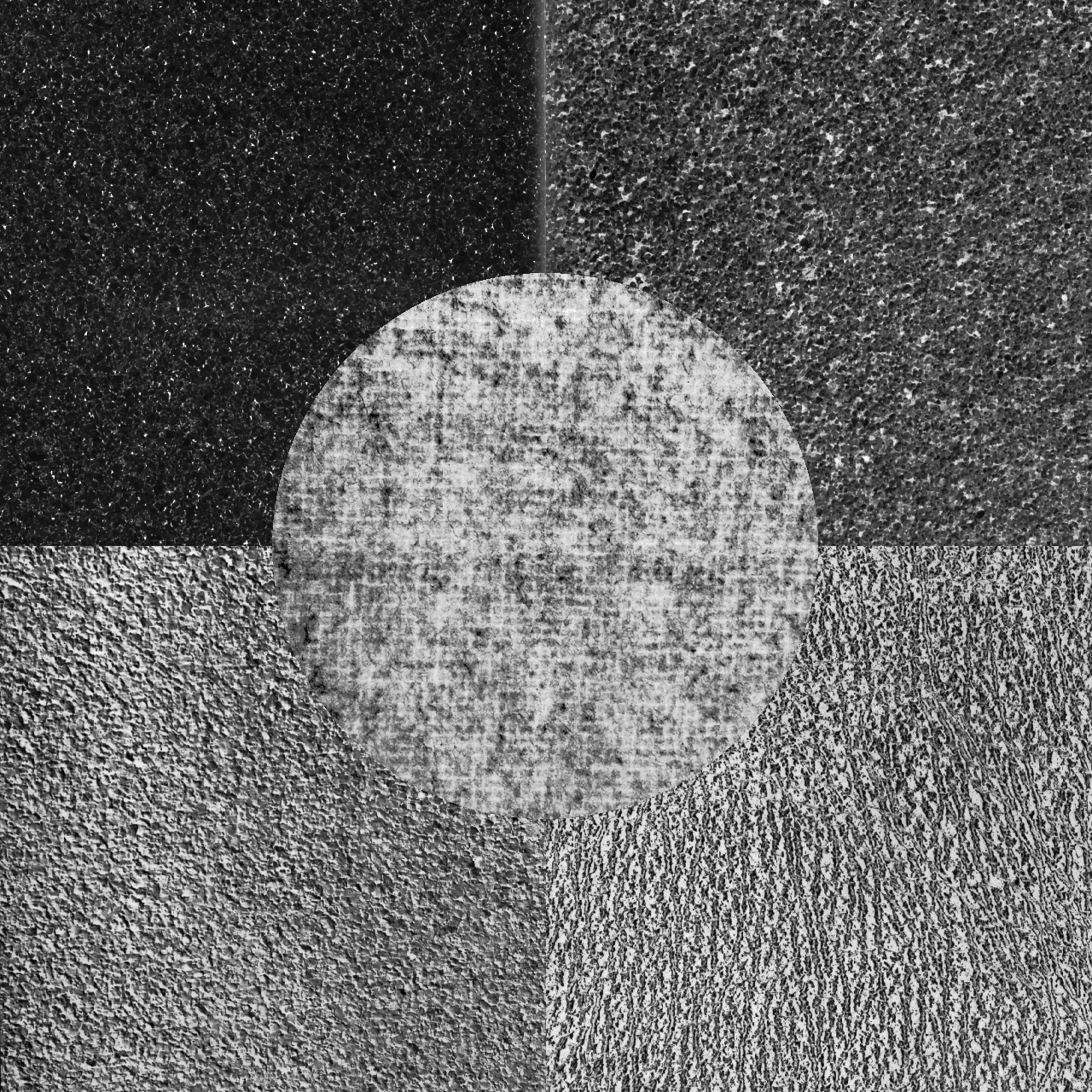}&
\includegraphics[scale=0.1]{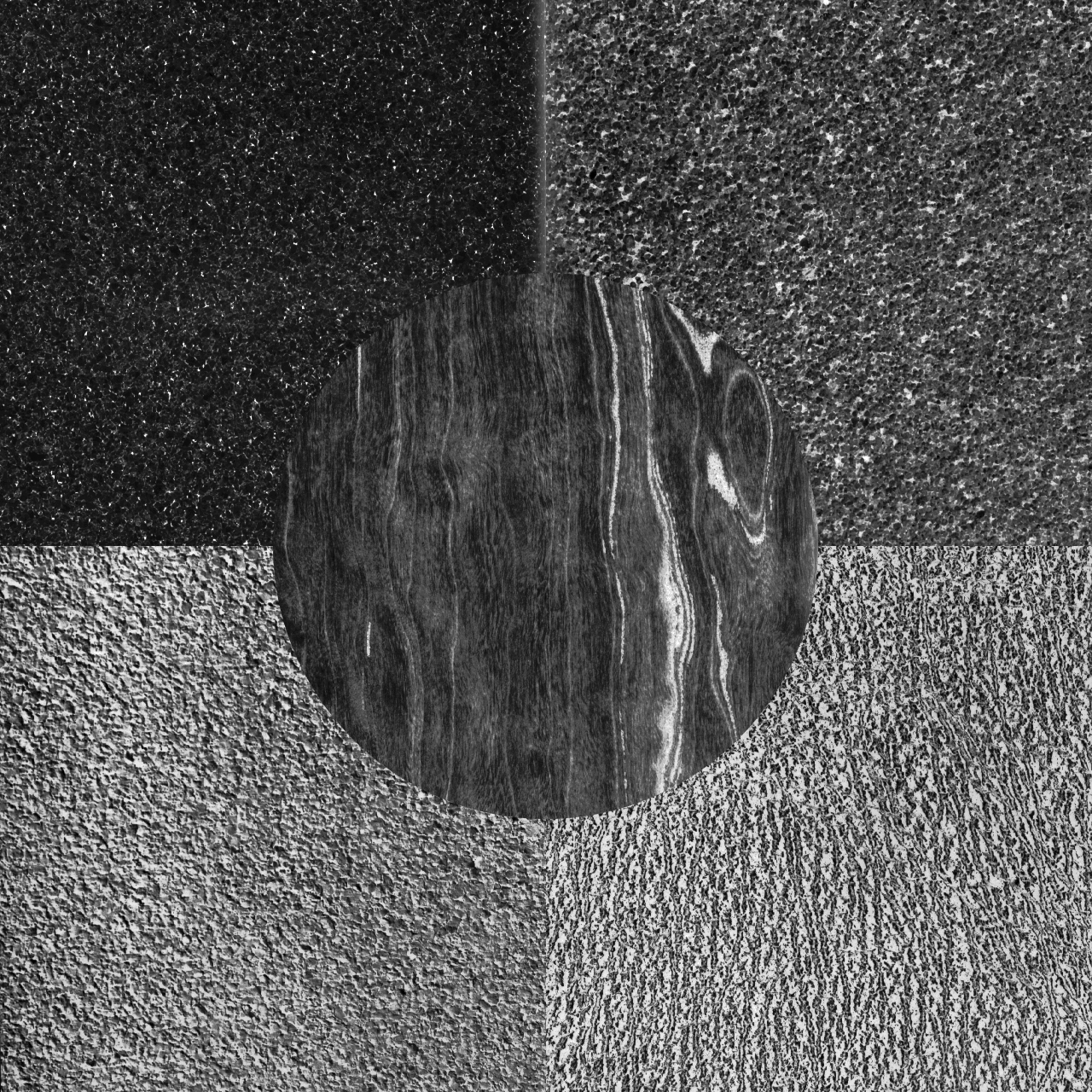}&
\includegraphics[scale=0.1]{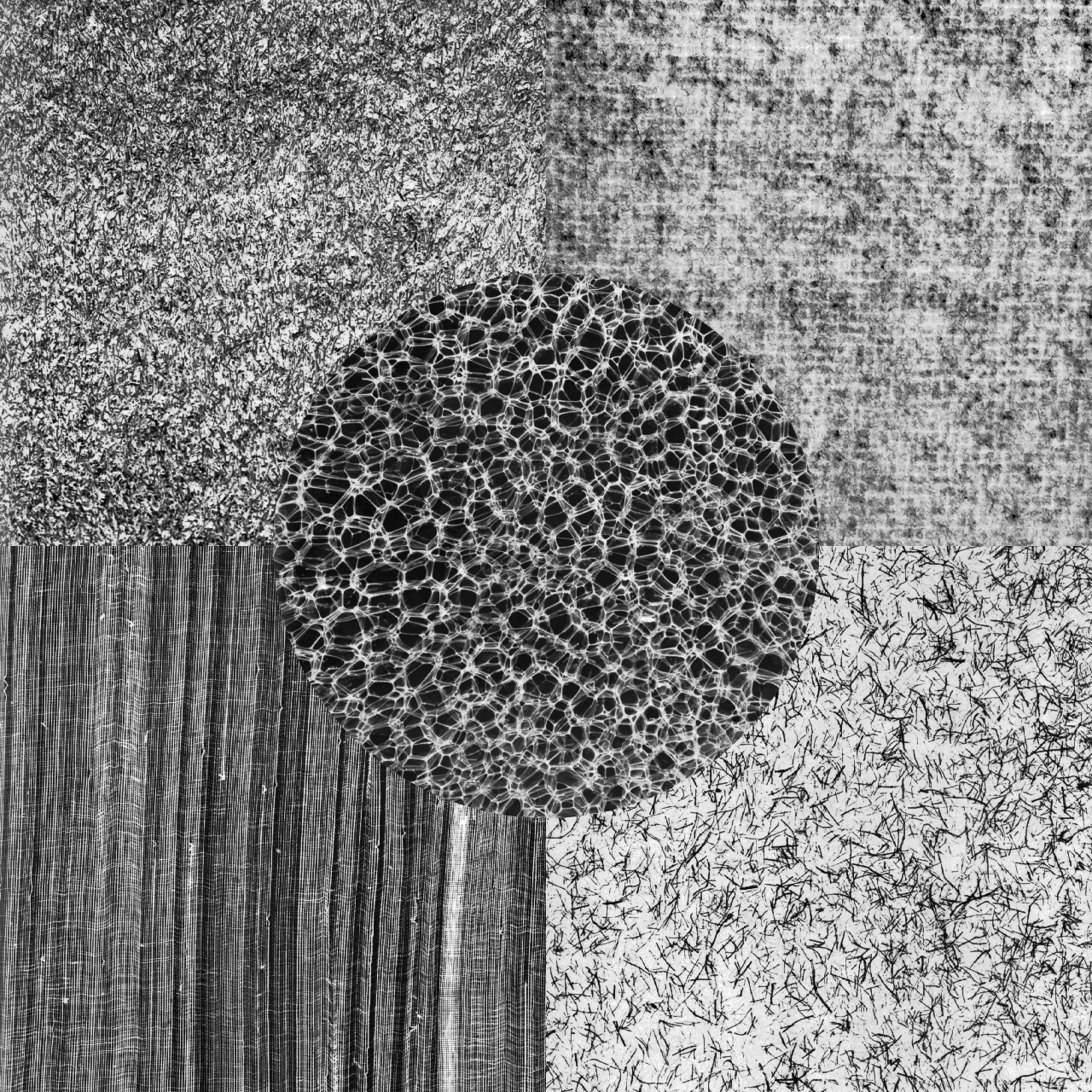}\\
(a)&(b)&(c)\\
\includegraphics[scale=0.1]{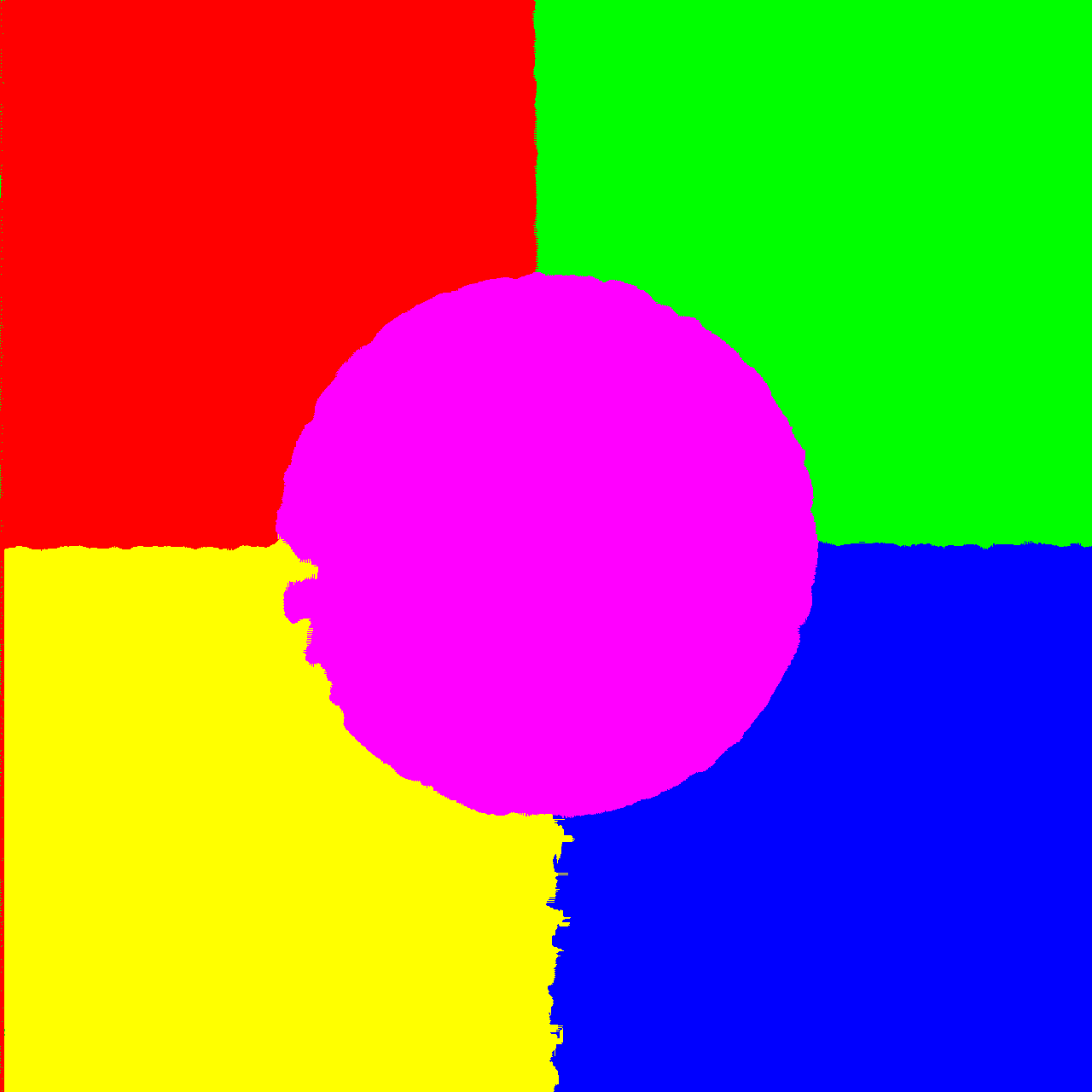}&
\includegraphics[scale=0.1]{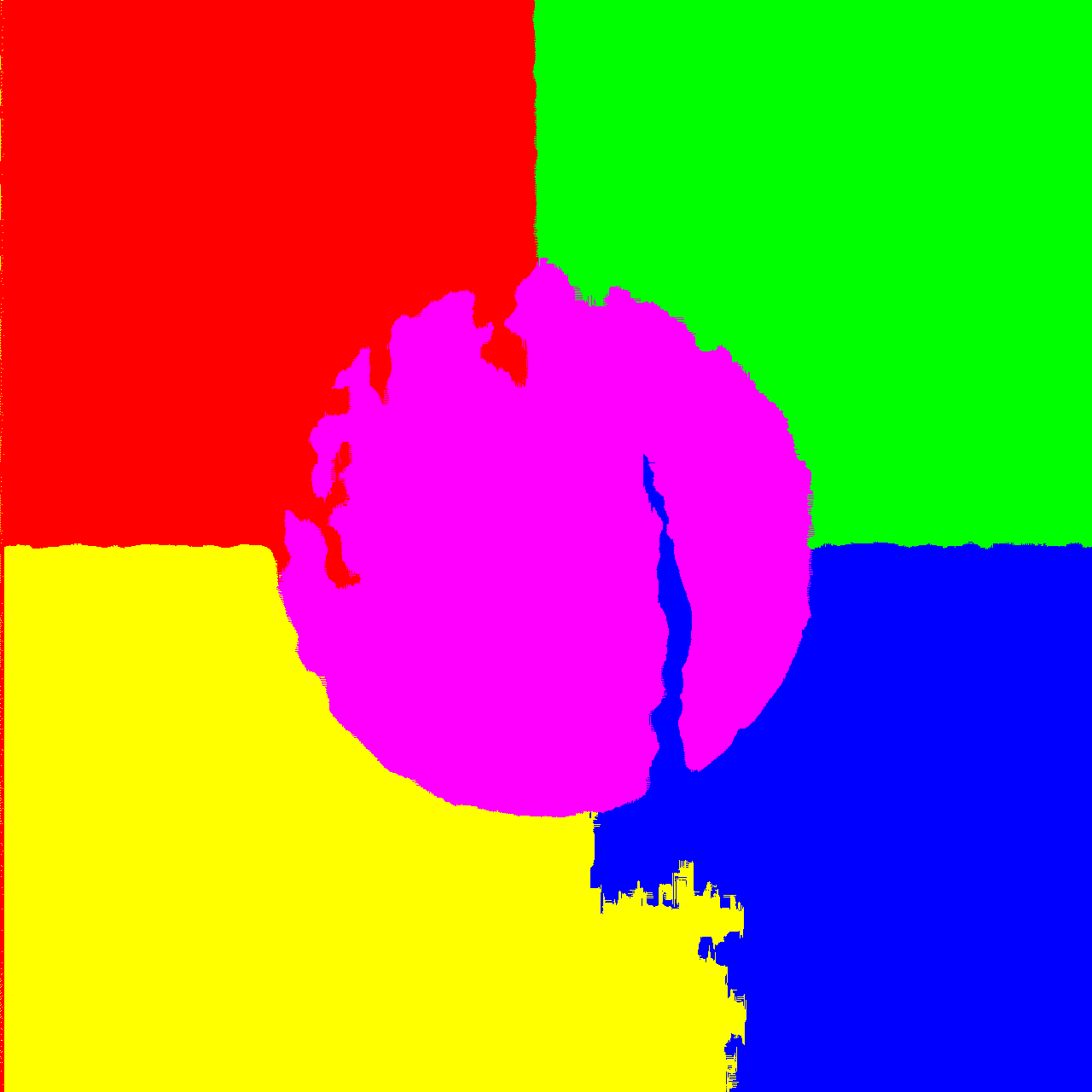}&
\includegraphics[scale=0.1]{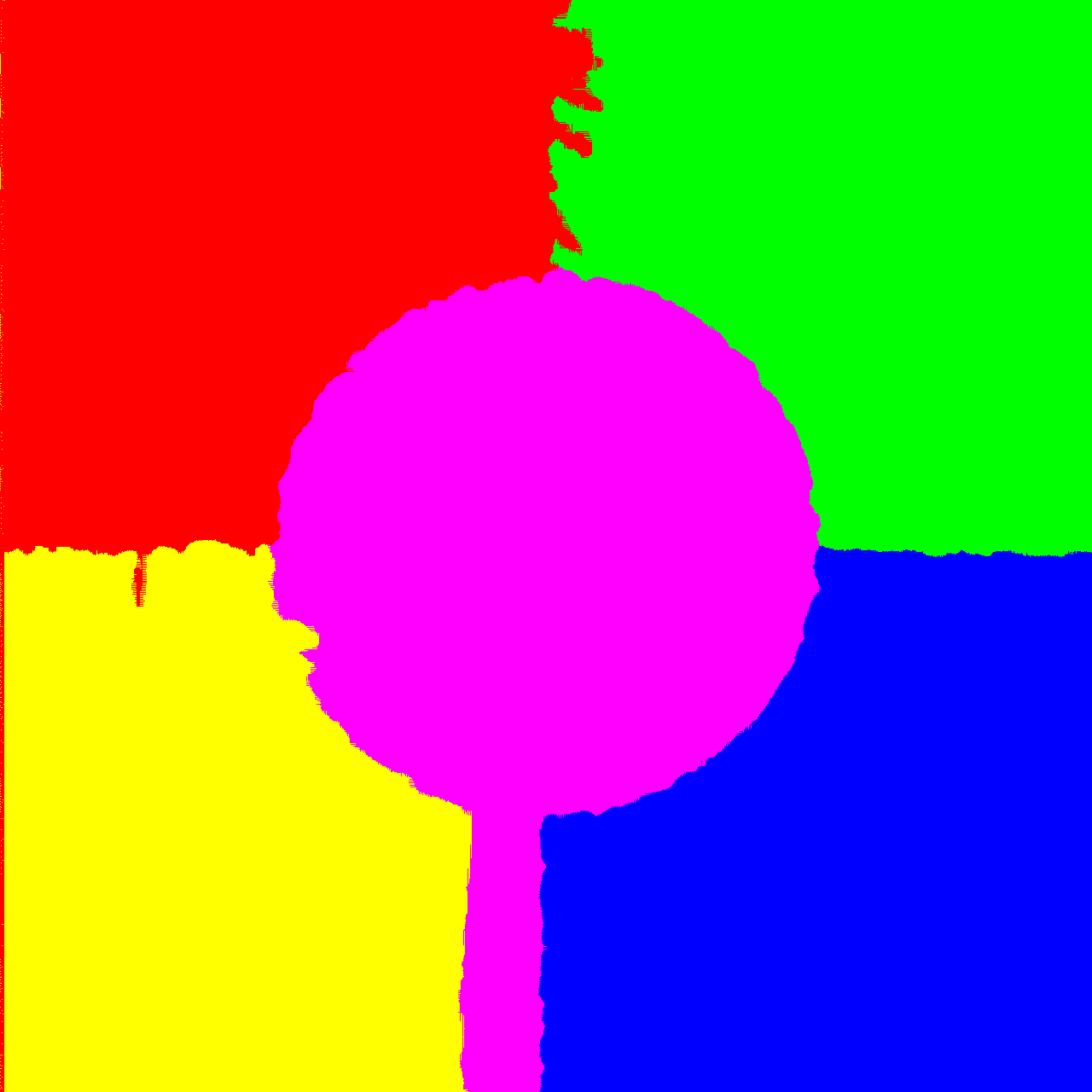}\\
(d)&(e)&(f)\\
\includegraphics[scale=0.1]{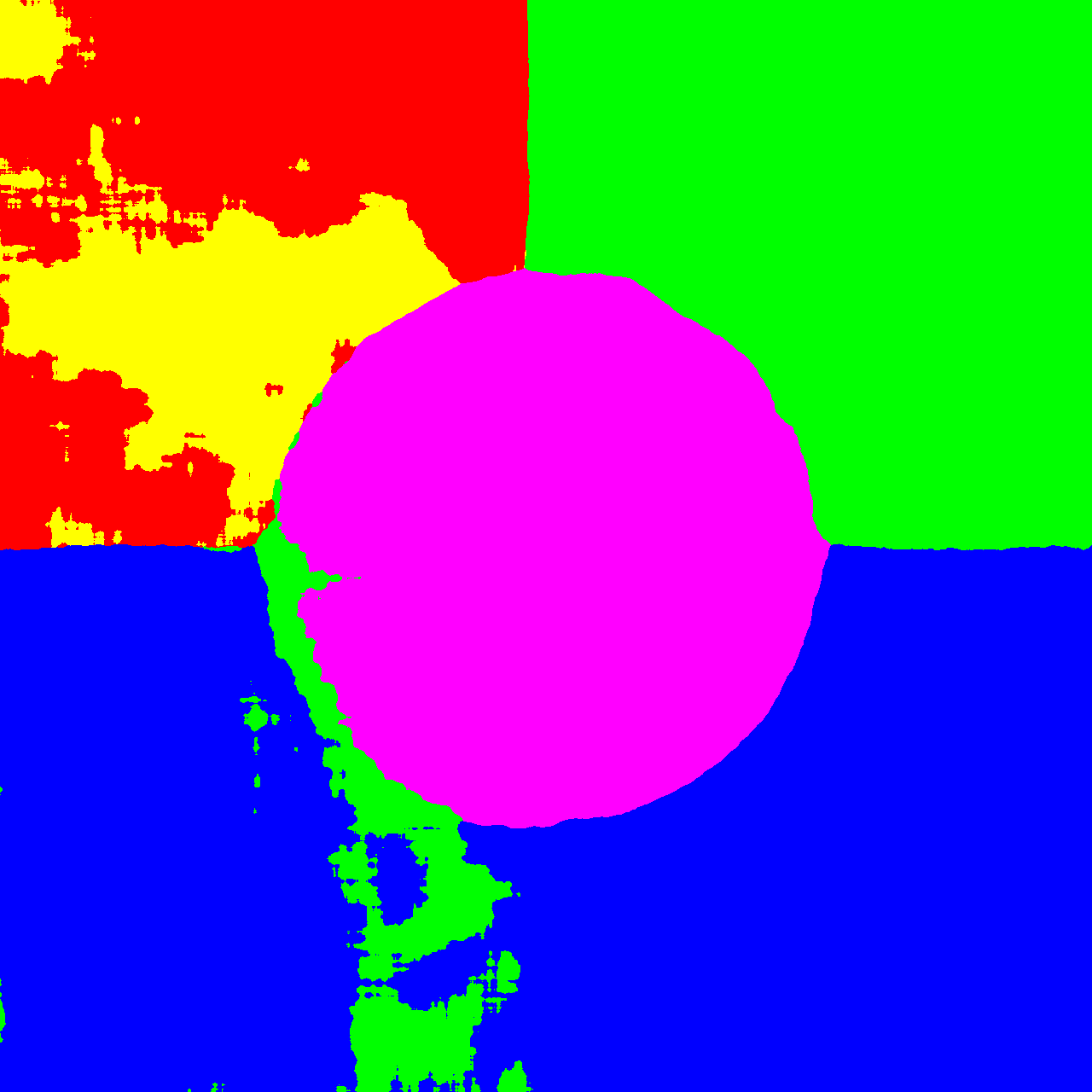}&
\includegraphics[scale=0.1]{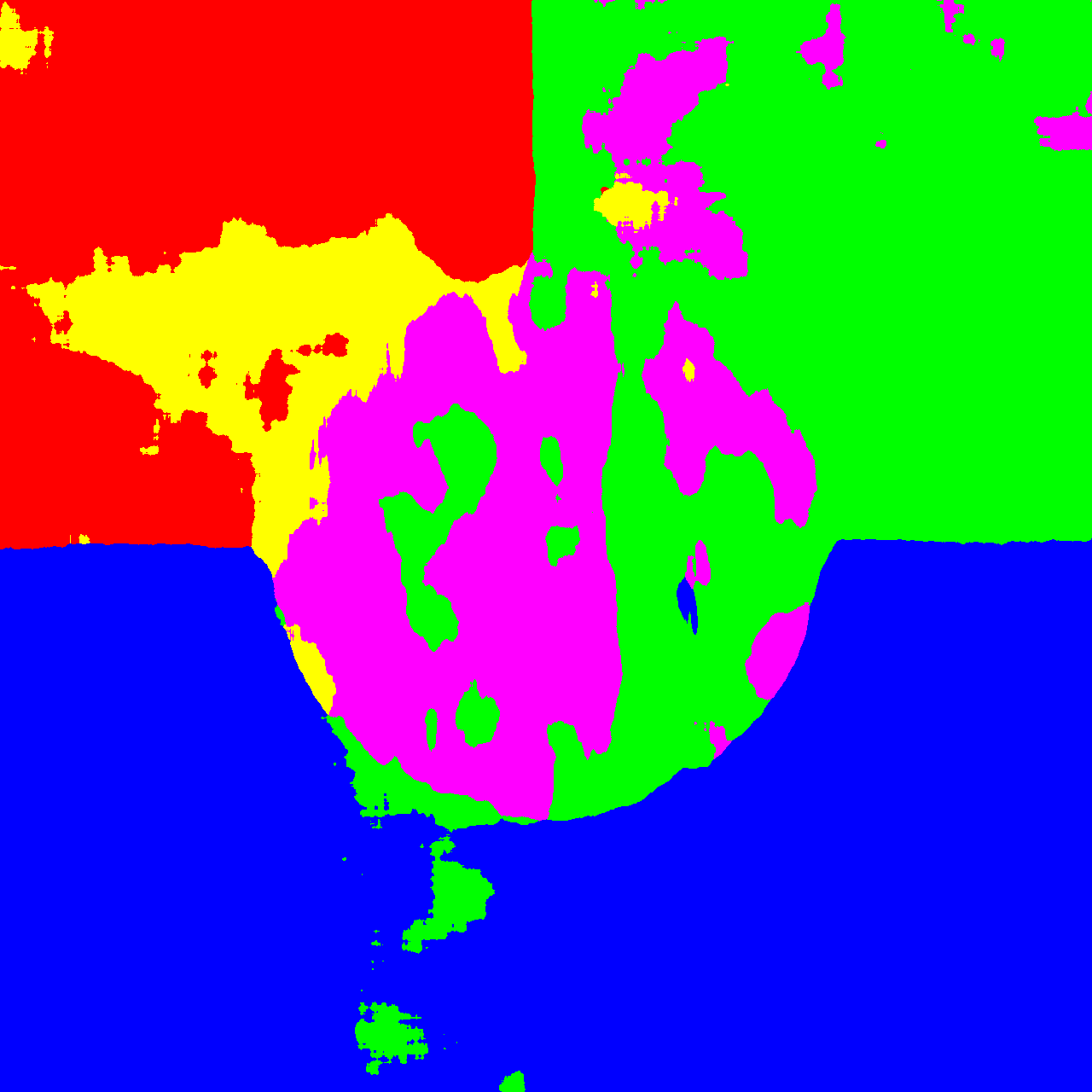}&
\includegraphics[scale=0.1]{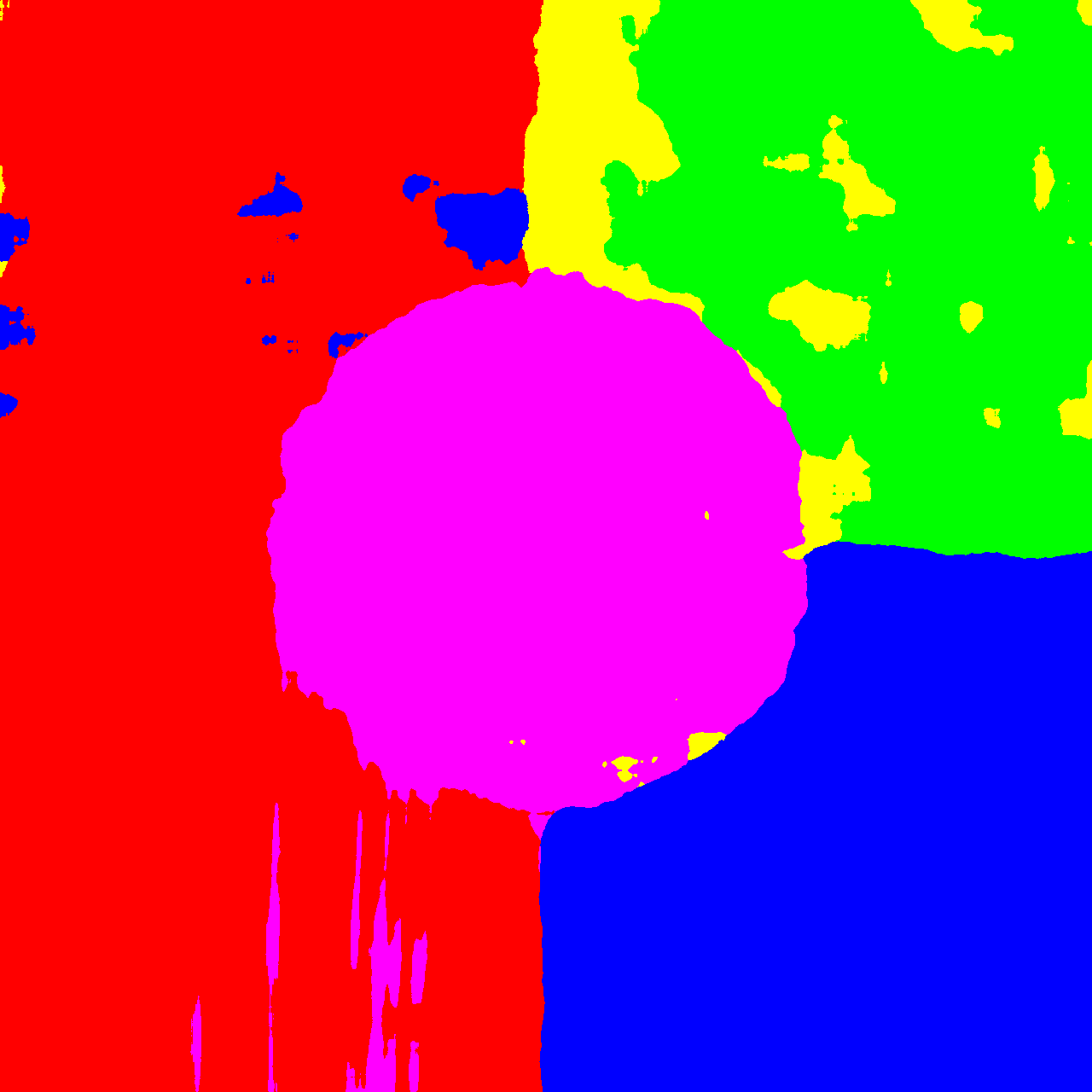}\\
(g)&(h)&(i)\\
\includegraphics[scale=0.1]{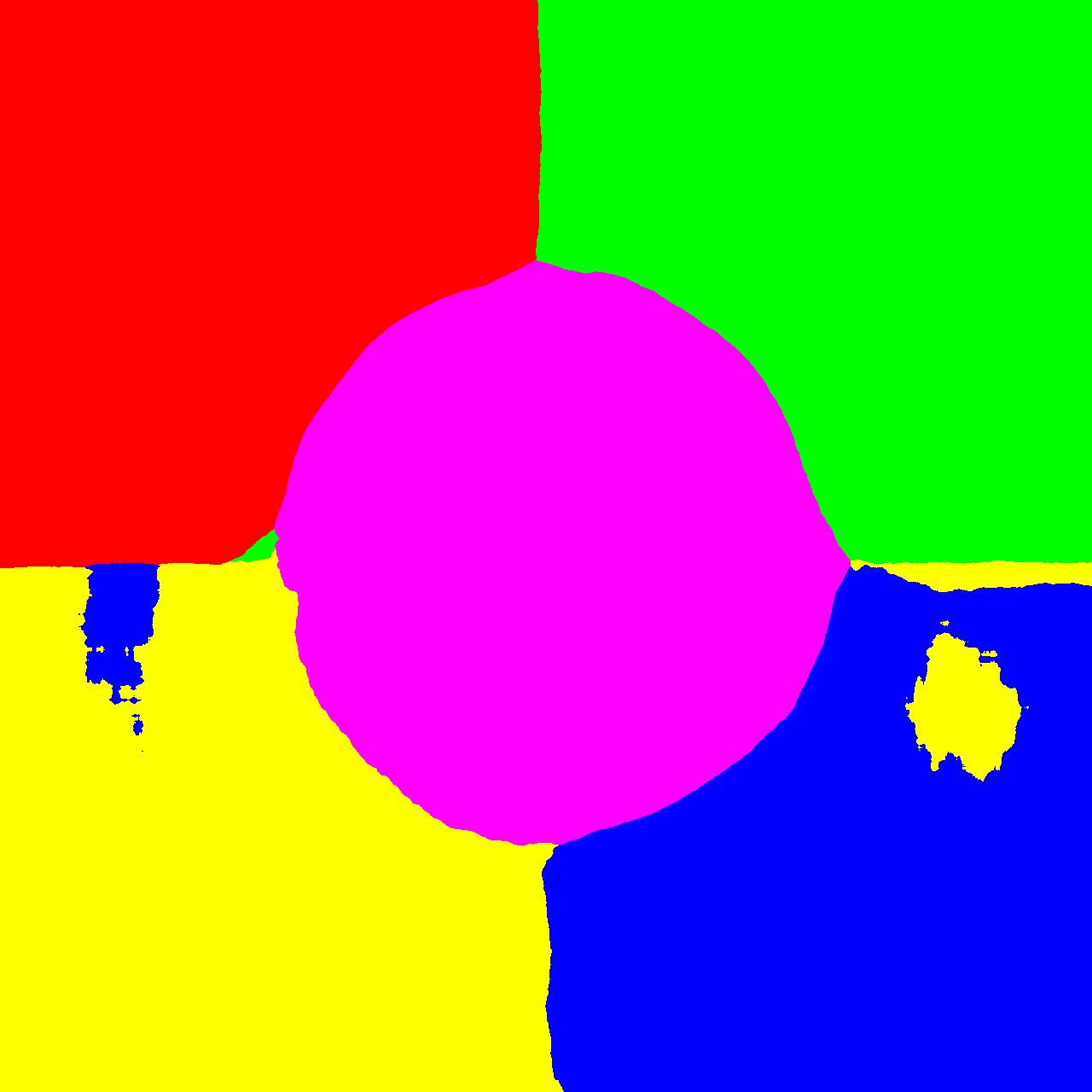}&
\includegraphics[scale=0.1]{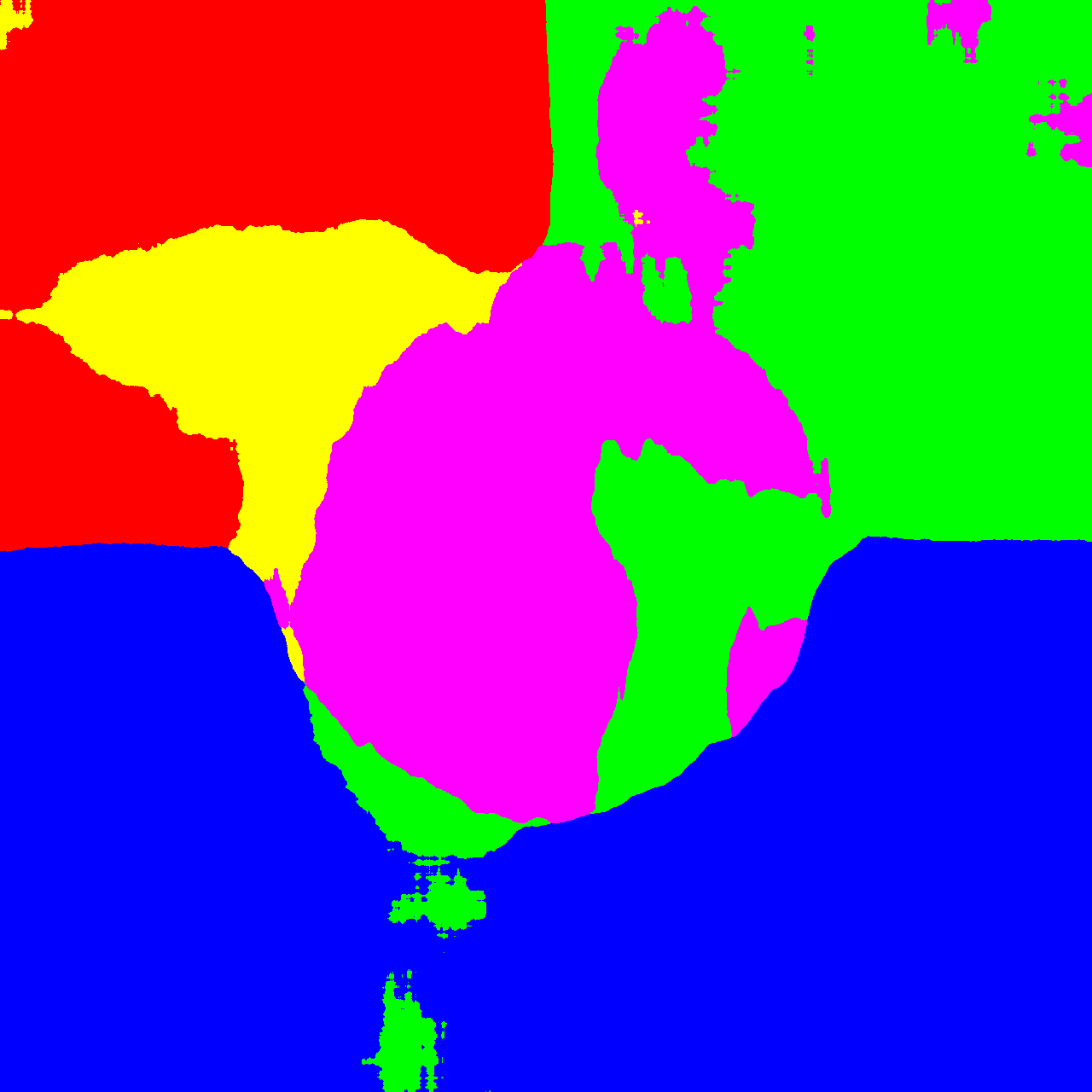}&
\includegraphics[scale=0.1]{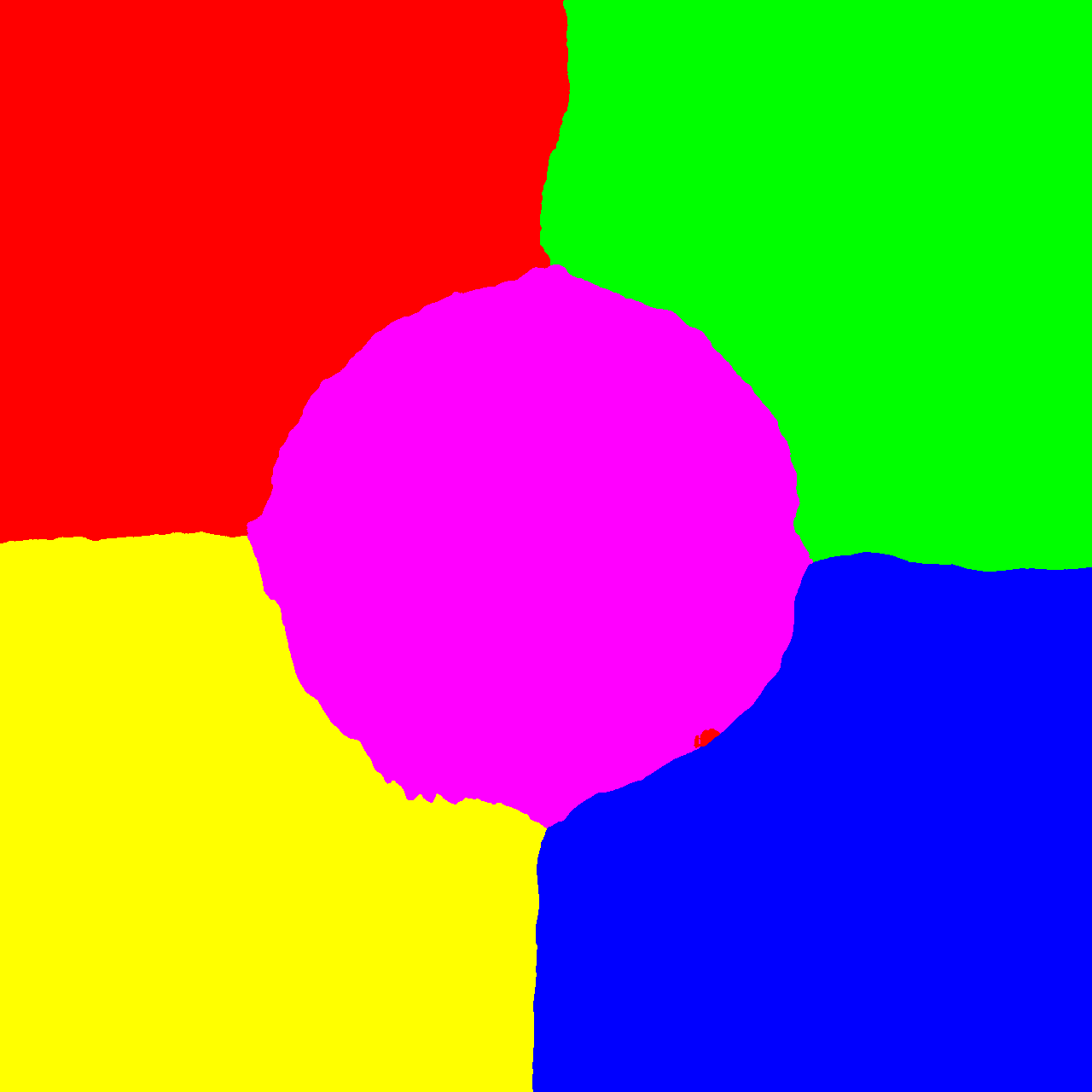}\\
(j)&(k)&(l)\\
\end{tabular}
\end{center}
\caption{Original mosaics produced by combining five Brodatz Textures (a), (b) and (c), and crisp segmentation maps for our proposed method (d), (e) and (f), for the method of \cite{YUAN15a} with integration scale of $60 \times 60$ (g), (h) and (i) and with integration scale of $60 \times 60$ (j), (k) and (l).}
\label{MosaicFiveTextures}
\end{figure*}

We tested the automatic seed selection algorithm on three mosaics of size $1280 \times 1280$ composed of five textures obtained from Brodatz Album \cite{BROD66a}. To illustrate the process, we show on Figure \ref{fig:cluster01} the clustering results for the mosaic shown on Figure \ref{MosaicFiveTextures}(a). The $k$-means clustering results show $5$ distinct sets of points, which means that selecting seeds near the centroid of each set may give us good seed candidates. After we choose the seeds for each class, we run the MOFS algorithm to segment the mosaic. In our experiments, the mosaics were divided in $64 \times 64$ patches with $20 \times 20$ pixels each. The statistics are then collected over the neighborhood whose size was computed as mentioned before.

Several recent papers have emerged in the area of image segmentation, especially using approaches based on training and deep learning \cite{SHEL17a, CIMP16a}. However, these works are proposed to segment images of generic characteristics, without any specialization for textural cases. Therefore we decided to compare the results obtained by our algorithm with the one proposed in \cite{YUAN15a} because, like our method, it proposes a method for segmenting images with textural properties without relying on any training.
The work of \cite{YUAN15a} presents a texture segmentation method based on a matrix factorization that is obtained through the use of pre-defined filters which act as texture descriptors for the image. Thus, the choice of these filters and the size of the integration scale (analogous to the size of neighborhood used in the MOFS) are determinant to the accuracy of the segmentation. In our experiments, we used the same filter bank proposed by Yuan et al. \cite{YUAN15a}, but we varied the integration scale between $60 \times 60$ (value used in the paper) and $120 \times 120$, for which we obtained the best segmentation results. 

\setlength{\tabcolsep}{3pt}

In Figure \ref{MosaicFiveTextures} we compare the results obtained using such configurations to the ones achieved by the automated version of MOFS using the two previously defined affinity functions and a maximum neighborhood window size of $15 \times 15$. In this figure, we show segmentation maps that do not code the fuzzy membership, for comparisons reasons. We computed Dice coefficients \cite{DICE45a} for each object and weighted them by the object sizes on the ground truth image, as suggested by \cite{CRUM06a}, producing a single score for each segmentation. The Dice coefficients and execution times for the methods are shown on Table \ref{AccuracyTable}. 

It is possible to notice that the increase of the integration scale (bottom row compared with fourth row of images on Figure \ref{MosaicFiveTextures}) tends to smooth the boundaries obtained by the segmentation, as well as to improve its accuracy. However, increasing it too much may lead to distorted segments that do not resemble the original texture's shape. Unlike our approach, for which we automatically chose an optimum window size to be used in affinity functions calculations, the method proposed by \cite{YUAN15a} was shown to be quite sensitive to the choice of the integration scale. In that sense, we ran their method as a semi-automatic one, by trying different scales and selecting the best one.

In addition to the fact that our method achieved better segmentation accuracy on three out of four cases, it also obtained boundaries that are closer to the original mosaics, an important requirement for various texture segmentation applications \cite{JOBA06a}. In the case where our method did not achieve the best segmentation accuracy, obtaining a Dice coefficient of $0.966$ compared to a Dice coefficient of $0.974$ of the \cite{YUAN15a} method, the region growing of the bottom left object was blocked by a white streak on that object. On a semi-automatic setting, the user would add another seed spel and execute the algorithm again. That is a flexibility that allows the method to be used on a general application setting.

Both algorithms have similar execution times - in the same order of magnitude. As MOFS segmentation of the mosaics took around $50s$ to finish, the method of \cite{YUAN15a} took about $30s$. Since there is a possibility of parallelization of the MOFS algorithm and considering that there is no requirement for executing this task in real time, such a time difference becomes insignificant.

\section{Conclusion}\label{sec:Conclusion}
In this paper we proposed a new method for using textural adaptive fuzzy spel affinities with the purpose of providing more robust texture segmentation. We employed two automatic methods for choosing the appropriate scale (neighborhood size) for a texture and for selecting seed spels for the objects, thus, turning the original semi-automatic method into an automatic one. We used fuzzy affinity functions that employ the skew divergence to compute the affinity between the histograms of neighbor windows. We have shown the successful usage of our method on several images that were composed by putting together several Brodatz textures. The comparison of the results produced by our method with the ones produced by the recently published method of \cite{YUAN15a} showed that our method was superior in three out of four cases, while generating more accurate boundaries.

Future work will include the application of this technique to low SNR applications, such as low-dosage Computerized Tomography (CT) exams, ultrasound (US) images, Synthetic Aperture Radar (SAR), optical microscopy rock sample images, and electron microscopy (EM) data. The usage of such technique to an specific application makes it possible to automate the whole process, by selecting the seed spels according to an appropriated heuristic, as shown in this article. We also plan to parallelize the MOFS algorithm, so it can be used to segment large three-dimensional volumes with textural properties in reasonable times.

{\small
\bibliographystyle{ieee}
\bibliography{TextureFuzzySeg}
}

\end{document}